\documentclass[letterpaper, 10 pt, journal, twoside]{IEEEtran}
\usepackage{times}
\usepackage[pdftex]{graphicx}
\usepackage{amsmath,amssymb,amsopn,amstext,amsfonts}
\usepackage{cancel}
\usepackage[space]{cite}
\usepackage{pdfsync}
\usepackage{balance}
\usepackage{color}
\usepackage{mathtools}
\usepackage{algpseudocode}
\usepackage{algorithm} 
\usepackage{bm}

\usepackage{diagbox}
\usepackage{float}
\usepackage{epstopdf}
\usepackage{pifont}
\usepackage{fixltx2e}
\usepackage{amsmath}
\usepackage{multirow}
\usepackage{url}
\usepackage{verbatim}
\usepackage{siunitx} 
\usepackage{epstopdf}

\usepackage[linkcolor=black,citecolor=black,urlcolor=black,colorlinks=true]{hyperref}
\usepackage{booktabs}
\usepackage{graphicx}
\usepackage{makecell}
\usepackage{enumerate}

\makeatletter
\let\NAT@parse\undefined
\makeatother
\usepackage{subfigure}
\usepackage{caption}
\hypersetup{draft}

\addcontentsline{toc}{section}{Acknowledgement}

\algnewcommand\algorithmicforeach{\textbf{for each}}
\algnewcommand{\algorithmicgoto}{\textbf{go to}}%
\algnewcommand{\Goto}[1]{\algorithmicgoto~\ref{#1}}%
\algdef{S}[FOR]{ForEach}[1]{\algorithmicforeach\ #1\ \algorithmicdo}

\usepackage{mathtools}

\newcommand{\Cov}{\boldsymbol{\Sigma}}

\newcommand{\rotvel}{\boldsymbol{\omega}}
\newcommand{\accvel}{\boldsymbol{a}}
\newcommand{\bias}{\boldsymbol{b}}
\newcommand{\bavel}{\bias_{\accvel}}
\newcommand{\bgvel}{\bias_{\rotvel}}
\newcommand{\navel}{\nb_{\accvel}}
\newcommand{\ngvel}{\nb_{\rotvel}}

\newcommand{\bodyframe}{\mathcal{B}}
\newcommand{\graframe}{\mathcal{G}}
\newcommand{\imuframe}{\mathcal{I}}

\DeclarePairedDelimiter\abs{\lvert}{\rvert}%
\DeclarePairedDelimiter\norm{\lVert}{\rVert}%

\makeatletter
\let\oldabs\abs
\def\abs{\@ifstar{\oldabs}{\oldabs*}}
\let\oldnorm\norm
\def\norm{\@ifstar{\oldnorm}{\oldnorm*}}
\makeatother

\usepackage{mathtools}

\input{macros/macro_alphabet.tex}

\graphicspath{{/figures/}}
\DeclareGraphicsExtensions{.png,.jpg,.eps,.pdf}
\IEEEoverridecommandlockouts

\title{DIDO: Deep Inertial Quadrotor Dynamical Odometry}

\author
{
	Kunyi Zhang$^{1,2}$, Chenxing Jiang$^{1,2}$, Jinghang Li$^{2}$, Sheng Yang$^{3}$, Teng Ma$^{3}$, Chao Xu$^{1,2}$, and Fei Gao$^{1,2}$
	\thanks{Manuscript received: February, 24, 2022; Revised: May, 19, 2022;
		Accepted: June, 17, 2022.}
	\thanks{
		This paper was recommended for publication by Editor Sven Behnke upon evaluation of the Associate Editor and Reviewers' comments.
		This work was supported by the National Key Research and Development Program of China (Grant NO. 2020AAA0108104), Alibaba Innovative Research (AIR) Program, and National natural Science Foundation of China under Grant 62003299.
		\big(\textit{Corresponding author: Fei Gao.}\big)}
	\thanks{\textsuperscript{1}State Key Laboratory of Industrial Control Technology, Institute of Cyber-Systems and Control, Zhejiang University, Hangzhou 310027, China.}
	\thanks{\textsuperscript{2}Huzhou Institute, Zhejiang University, Huzhou 313000, China.}
	\thanks{\textsuperscript{3}Alibaba DAMO Academy Autonomous Driving Lab, Hangzhou 311121, China.}
	\thanks{E-mail: {\{kunyizhang, fgaoaa\}@zju.edu.cn}}
	\thanks{Digital Object Identifier (DOI): see top of this page.}
}

\begin{document}
	
	\maketitle
	
	\begin{abstract}   
		
		In this work, we propose an interoceptive-only state estimation system for a quadrotor with deep neural network processing, where the quadrotor dynamics is considered as a perceptive supplement of the inertial kinematics.
		To improve the precision of multi-sensor fusion, we train cascaded networks on real-world quadrotor flight data to learn IMU kinematic properties, quadrotor dynamic characteristics, and motion states of the quadrotor along with their uncertainty information, respectively. 
		This encoded information empowers us to address the issues of IMU bias stability, quadrotor dynamics, and multi-sensor calibration during sensor fusion. 
		The above multi-source information is fused into a two-stage Extended Kalman Filter (EKF) framework for better estimation. 
		Experiments have demonstrated the advantages of our proposed work over several conventional and learning-based methods.
		
		%
	\end{abstract}

	

	\section{Introduction}
	\IEEEPARstart{A}{erial} robots are popular autonomous vehicles prevalently used in entertainment and industrial applications. To locate themselves, exteroceptive sensors (e.g., LiDAR and cameras) play a prominent role in common scenarios. However, when encountering degenerate cases where features are either heavily repeated or insufficient, these robots in-turn rely on interoceptive sensors (e.g., IMU and rotor tachometer) to deduct their poses. 
	Inaccurate dynamics models, unestimated bias can lead to rapid divergence of state propagation.
	
	Recent approaches~\cite{chen2018ionet,yan2019ronin,yan2018ridi,brossard2019ai,sun2021idol,liu2020tlio}, for exteroceptive sensor disabled scenarios, choose to learn the correspondence between IMU sequences and the relative poses through a data-driven approach.
	Although these works show the potential of deep neural networks for estimation, they are typically designed for scenes of walking pedestrians or low-degree-of-freedom (DOF) ground vehicles, which is difficult to generalize to aerial robots.
	
	As a 4 DOF under-actuated system, there is a coupling of kinematic states between the velocity and tilt of the quadrotor, which are not fully exploited in the general state estimation system.
	The algorithm proposed in~\cite{svacha2019inertial} can obtain the tilt and velocity of a quadrotor using only the tachometer and IMU data.
	However, its dynamics model only considers a quadratic thrust model and a first-order drag model, which the quadrotors do not strictly conform to.
	NeuroBEM~\cite{bauersfeld2021neurobem} employs a neural network to compensate for the unmodeled parts such as airflow during the dynamics modeling, but it is not utilized for state estimation.
	\begin{figure}[t!]
		\vspace{0.1cm}
		\centering
		\includegraphics[width=1\linewidth]{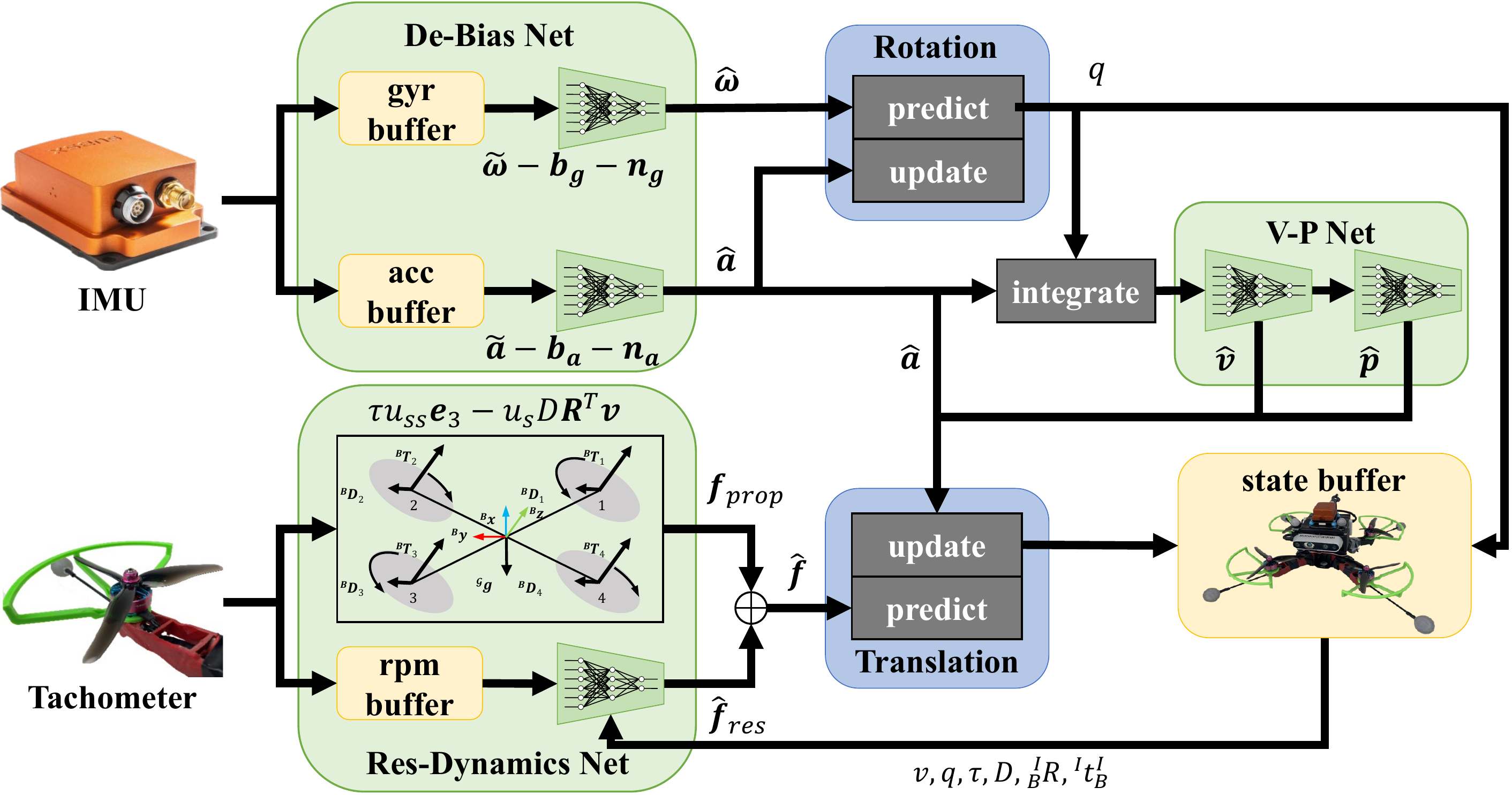}
		\caption{
			The filter pipeline of proposed Deep Inertial quadrotor Dynamical Odometry~(DIDO). The green parts are cascaded networks designed for learning kinematic properties, quadrotor dynamic characteristics, and motion states. The blue parts are EKF processes.
		}
		\label{fig:overview}
		\vspace{-1.0cm}
	\end{figure}
	
	Therefore, this paper proposes a deep inertial-dynamical odometry framework, employing neural networks to learn both IMU and dynamics properties for better performance in quadrotor state estimation. 
	The kinematic properties of IMU (accelerometer and gyroscope) are learned by two convolutional neural networks~(CNN) for measurement de-biasing.
	The dynamic characteristics of quadrotors, with respect to the propeller rotation speed and states, are learned to compensate for mechanistic modeling by a CNN. 
	Besides, the velocity and position of the quadrotor are observed in real-time by a recurrent neural network~(RNN).
	Finally, the networks mentioned above are fused with IMU and tachometer measurements in a two-stage EKF framework.
	Specifically, in the stage of rotation, the de-biased angular velocity and acceleration are fused to get accurate rotation by gravity-alignment.
	In the translation stage, the modified dynamics is regard as process model, and the acceleration along with velocity and position from the network are considered as observation models, for accurate estimation of kinematic, dynamic states, and extrinsic parameters. 
	The letter's contributions are enumerated below:
	\begin{enumerate}
		\item We present a series of networks to learn the IMU kinematics and quadrotor dynamics, demonstrating the capability to provide precise observations with only raw IMU and tachometer data.
		\item We propose a complete state estimation system that combines quadrotor dynamics and network observations within a two-stage EKF to jointly estimate kinematic, dynamic states, and extrinsic parameters.
	\end{enumerate}
	
	
	\section{Related Works}
	\label{sec:Related Works}
	%
	
	\subsection{IMU Learning based estimation}
	IONet~\cite{chen2018ionet}, RoNIN~\cite{yan2019ronin} and RIDI~\cite{yan2018ridi} are end-to-end learning methods for pose estimation.
	IONet~\cite{chen2018ionet} uses a long short-term memory network~(LSTM) based network, whose inputs are accelerometer and gyroscope measurements within a time window in the world coordinate system.
	RoNIN~\cite{yan2019ronin} proposes three different neural network architectures to solve the inertial navigation problem: LSTM, temporal convolutional network~(TCN), and residual network~(ResNet). 
	These models regress the velocity and displacement of the pedestrian in 2 dimensions.
	RIDI~\cite{yan2018ridi} design a two-stage system, which first regresses the velocity of pedestrians and then optimizes correction terms for IMU to match the velocity. The displacement of a pedestrian is finally obtained after integration. 
	
	AI-IMU~\cite{brossard2019ai}, IDOL~\cite{sun2021idol}, TLIO~\cite{liu2020tlio} and CTIN~\cite{rao2022ctin}
	 combines deep neural networks with filter framework to obtain better performance.
	AI-IMU~\cite{brossard2019ai} is specially designed for the purpose of vehicle's pose estimation, using a CNN to estimate the covariance of accelerometer and gyroscope measurements and applying them to an invariant extended Kalman filter (IEKF).
	In addition, the filter considers the vehicle sideslip constraints as well as the motion constraints in the vertical direction.
	IDOL~\cite{sun2021idol} and TLIO~\cite{liu2020tlio} are both applied for pedestrian's pose estimation.
	IDOL~\cite{sun2021idol} is divided into a rotation estimation module and a translation estimation module.
	In the rotation module, an LSTM network with gyroscope, accelerometer, and magnetometer data as inputs is trained to output the rotation, which is fused with the rotation obtained by integrating the gyroscope using the extended Kalman filter (EKF). 
	In the position estimation module, the position is regressed using the LSTM network after converting the acceleration to the world coordinate system using the above rotation.
	TLIO~\cite{liu2020tlio} uses a ResNet network similar to RoNIN~\cite{yan2019ronin} to estimate pedestrian displacements without yaw, and a stochastic cloning extended Kalman filter (SCEKF)~\cite{roumeliotis2002stochastic} framework to fuse the output values of the network with raw IMU measurements, for estimating position, rotation, and sensor bias in a tightly coupled manner.
	CTIN~\cite{rao2022ctin} uses ResNet and LSTM to obtain the local and global embeddings, and feeds these two embeddings into two multi-headed Transformers~\cite{vaswani2017attention} to obtain the velocity and its covariance.
	
	\subsection{Quadrotor dynamics and state estimation}
	A quadrotor is usually powered by four rotating propellers, and the relationship between rotors' speed and dynamical states can be obtained by modeling the quadrotor's mechanics.
	Traditionally, a simple quadratic model is widely adopted, where the thrust and axial torque generated by the rotor are proportional to the square of their rotation speed with a constant factor~\cite{mahony2012multirotor}. The momentum theorem does not take into account either the motion of the rotor in air or the interactions between these rotors and the body.
	The blade-element-momentum (BEM) theory~\cite{khan2013toward, gill2017propeller,gill2019computationally} combines blade-element theory and momentum theory to alleviate the difficulty of calculating the induced velocity and accurately capture the aerodynamic force and torque acting on single rotor in a wide range of operating conditions. But it also does not take into account any interactions between different propellers and body.
	
	NeuroBEM~\cite{bauersfeld2021neurobem} uses a neural network to fit the residuals between the real quadrotor dynamics and the calculated values from the BEM theory. It~\cite{bauersfeld2021neurobem} takes the rotor speed and kinematic states as the network's inputs to allow the quadrotor to perform highly maneuverable trajectories.
	In~\cite{svacha2019inertial}, the authors estimate the tilt and velocity of the quadrotor in the body frame by applying only the tachometer and gyroscope as inputs and taking the accelerometer as observation.
	VIMO~\cite{nisar2019vimo} and VID-Fusion~\cite{ding2020vid} also consider the constraints of quadrotor dynamics in visual inertial fusion to improve the robustness and accuracy of pose estimation.
	
	\section{Kinematic and Dynamic Network Design}
	\label{sec:Network}
	\subsection{Preliminaries} 
	\subsubsection{IMU Model} 
	\label{sec:imu}
	IMU measurements include the gyroscope $\widetilde{\rotvel}$ and non-gravitational acceleration $\widetilde{\accvel}$, which are measured in the IMU frame~(the $\imuframe$ frame with IMU sensor as the center, Front-Left-Up as the order of three axes) and given by:
	\begin{equation}
	\begin{aligned}
	{^{\imuframe}}\widetilde{\rotvel} &= 
	{^{\imuframe}}{\rotvel} + {^{\imuframe}}{\bgvel} + {\ngvel}, \\
	{^{\imuframe}}\widetilde{\accvel} &= 
	{^{\imuframe}}{\accvel} + {^{\imuframe}}{\bavel} + {_{\imuframe}^{\graframe}}\mathbf{R}^\top {^{\graframe}}\boldsymbol{\mathrm{g}} + {\navel},
	\end{aligned}
	\end{equation}
	where
	${^{\imuframe}}{\rotvel}$ and ${^{\imuframe}}{\accvel}$ are the true angular velocity and acceleration,
	${^{\graframe}}\boldsymbol{\mathrm{g}} = [0,0,9.8]$ (unit: $m/s^2$) is the gravity vector in the gravity-aligned frame~(the ${\graframe}$ frame with z-axis pointing down vertically),
	${_{\imuframe}^{\graframe}}\mathbf{R}$ is the rotation matrix from the $\imuframe$ frame to the ${\graframe}$ frame,
	$\nb_{\rotvel}$ and ${\navel}$ are the additive Gaussian white noise in gyroscope and acceleration measurements,
	${\bgvel}$ and ${\bavel}$ are the bias of IMU modeled as random walk:
	\begin{equation}
	\begin{aligned}
	{\ngvel} &\sim \mathcal{N}(0,\Cov_{\rotvel}^2), &
	\dot{\bias}_{\rotvel} &\sim \mathcal{N}(0,\Cov_{\bgvel}^2), \\
	{\navel} &\sim \mathcal{N}(0,\Cov_{\accvel}^2), &
	\dot{\bias}_{\accvel} &\sim \mathcal{N}(0,\Cov_{\bavel}^2).
	\end{aligned}
	\end{equation}
	
	\subsubsection{Quadrotor Dynamics}
	\label{dynamics}
	Because the propagation of kinematic states is driven by multiple propulsion units in a quadrotor system, we model the Newtonian dynamics according to~\cite{svacha2020imu}.
	The total driving force of a quadrotor in the body frame~(the $\bodyframe$ frame with center of mass as the center, the same order as the $\imuframe$ frame) is the sum of the thrust ${^\bodyframe\boldsymbol{F}_t}$ and drag force ${^\bodyframe\boldsymbol{F}_d}$ generated by each propulsion unit as follow:
	\begin{equation}
	\label{dynamics_equation}
	\begin{aligned}
	^\bodyframe\boldsymbol{F}
	=\sum_{i=1}^{4} \left( {^\bodyframe\boldsymbol{F}}_{t_i}  
	-{^\bodyframe\boldsymbol{F}}_{d_i} \right) 
	=\sum_{i=1}^{4} \left( \tau u_{i}^{2} \boldsymbol{e}_3
	-u_{i} D {^\bodyframe\boldsymbol{v}}_i \right),
	\end{aligned} 
	\end{equation}
	where
	$\tau$ is the thrust coefficient for the propellers, 
	$D=diag(d_x,d_y,d_z)$ is the matrix of effective linear drag coefficients,
	$\boldsymbol{e}_3 = [0,0,1]^\top$ is the $z$ axis in any frame,
	and $u_{i}$ and ${^\bodyframe\boldsymbol{v}}_i$ are the rotation speed and velocity of the $i$-th rotor, respectively.
	Actually, the velocity of each rotor is 
	${^\bodyframe}\boldsymbol{v}_i={^\bodyframe}\boldsymbol{v}+{^\bodyframe}{\rotvel} \times{^\bodyframe}\boldsymbol{r}{_i^{\bodyframe}}$,
	where
	${^\bodyframe}\boldsymbol{v}$ and ${^\bodyframe}{\rotvel}$ are the linear and angular velocity of the quadrotor's center of mass (COM),
	${^\bodyframe}\boldsymbol{r}{_i^{\bodyframe}}$ is the position of the $i$-th rotor relative to the COM. To simplify the calculation, we ignore the velocity discrepancy of different rotors and express it as ${^\bodyframe}\boldsymbol{v}_i \approx {^\bodyframe}\boldsymbol{v}$.
	The input notations are abbreviated as $U_{ss} = \sum_{i=1}^{4} u_{i}^{2}$ and $U_{s} = \sum_{i=1}^{4} u_{i}$.
	Therefore, we can obtain the Newtonian equation in the ${\graframe}$ frame:
	\begin{equation}
	\label{newtonian_equation}
	\begin{aligned}
	m\frac{d}{dt} \left( {{^{\graframe}}\boldsymbol{v}}  \right) 
	&= {_{\bodyframe}^{\graframe}}\mathbf{R} 
	\left(
	\tau U_{ss} {\boldsymbol{e}}_3 - 
	U_{s} D {_{\bodyframe}^{\graframe}}\mathbf{R}^\top 
	{{^{\graframe}}\boldsymbol{v}{_{\bodyframe}^{\graframe}}}
	\right)
	- m{^{\graframe}}\boldsymbol{\mathrm{g}}. \\
	\end{aligned} 
	\end{equation}
	Note that, we divide the rotor speed by $10,000$ to ensure the numerical stability.

	\subsection{Network Design}
	\subsubsection{De-Bias Net}
	\label{debias}
	As an interoceptive sensor, IMU often requires several integrations to obtain the kinematic states, 
	but the system would drift or diverge because of the failure to accurately estimate the accelerometer bias ${^{\imuframe}}\bias_a$ and gyroscope bias ${^{\imuframe}}\bias_\omega$.
	So, we design a~\textit{De-Bias Net} to learn the kinematic characteristics of IMU for de-biasing.
	The two \textit{De-Bias Net}s for accelerometer and gyroscope have the same network architecture, 
	a 1D version of ResNet~\cite{he2016deep} with only one residual block to boost the inference process.
	Fully-connected layers are extended to output.
	
	The respective input features of both~\textit{De-Bias Net}s are historical raw accelerometer measurements ${^{\imuframe}}\widetilde{\accvel}$ and gyroscope measurements ${^{\imuframe}} \widetilde{\rotvel}$,
	each output is ${^{\imuframe}}\widehat{\bias}_{\accvel}$ and ${^{\imuframe}}\widehat{\bias}_{\rotvel}$ at every moment, separately. 
	
	We define two Mean Square Error (MSE) loss functions $\mathcal{L}_{\text{MSE},{\accvel}}$ and $\mathcal{L}_{\text{MSE},{\rotvel}}$ for accelerometer and gyroscope, respectively, on the following integrated increments:
	\begin{equation}
	\begin{aligned}
	\mathcal{L}_{\text{MSE},{\accvel}} &= 
	\frac{1}{N}\sum_{i=1}^N
	{\rVert 
		{\boldsymbol v}_{i,i+n} - 
		{\boldsymbol{\widehat v}}_{i,i+n}
		\rVert^2_2}, \\ 
	\mathcal{L}_{\text{MSE},{\rotvel}} &= 
	\frac{1}{N}\sum_{i=1}^N
	{\Vert 
		\log \left(
		\left(
		{\boldsymbol{\widehat q}}_{i,i+n} \right)^* \otimes 
		{\boldsymbol q}_{i,i+n} \right)
		\Vert^2_2},
	\end{aligned}
	\end{equation}
	where
	\begin{equation}
	\begin{aligned}
	{\boldsymbol{\widehat v}}_{i,i+n}
	&= \int_{i}^{i+n} 
	{_{\imuframe_t}^{\graframe}}\mathbf{R} 
	({^{\imuframe_t}}\widetilde{\accvel} - {^{\imuframe_t}}\widehat{\bias}_{\accvel}) dt, \\
	{\boldsymbol{v}}_{i,i+n}
	&= {^{\graframe}}\boldsymbol{v}_{i+n}-{^{\graframe}}\boldsymbol{v}_{i}, \\
	{\boldsymbol{\widehat q}}_{i,i+n}
	&= \int_{i}^{i+n}  
	\frac{1}{2} {_{\imuframe_t}^{\imuframe_{i}}} {\boldsymbol{q}} \otimes 
	({^{\imuframe_t}}\widetilde{\rotvel} - 
	{^{\imuframe_t}}\widehat{\bias}_{\rotvel}) dt,\\
	{{\boldsymbol q}}_{i,i+n}
	&= ({_{\imuframe_{i}}^{\graframe}} {\boldsymbol{q}})^* \otimes 
	{_{\imuframe_{i+n}}^{\graframe}} {\boldsymbol{q}}.
	\end{aligned}
	\end{equation}
	Note that, the subscript $i$ is the abbreviation for time $t_i$,
	${\boldsymbol{\widehat v}}_{i,i+n}$ and ${\boldsymbol{\widehat q}}_{i,i+n}$ are the intermediate variables calculated from the estimated values
	${^{\imuframe}}\widehat{\bias}_{\accvel}$ and ${^{\imuframe}}\widehat{\bias}_{\rotvel}$ of the~\textit{De-Bias Net} by the forward Euler method actually,
	${\boldsymbol{v}}_{i,i+n}$ and ${{\boldsymbol q}}_{i,i+n}$ are the ground truth velocity increment and relative rotation of each sliding window, $n$ is the window size, $N$ is the batch size, and $\otimes$ means the quaternion multiplication. In training, ${_{\imuframe}^{\graframe}}\mathbf{R}$ is the ground truth rotation.
	
	\subsubsection{Res-Dynamics Net} 
	Actually, the dynamics model of quadratic thrust and approximately linear drag in Eq.~(\ref{dynamics_equation}) can not accurately model quadrotors because of complicated effects related to airflow and so on.
	Therefore, like NeuroBEM~\cite{bauersfeld2021neurobem}, we adopt the same ResNet~\cite{he2016deep} as~\textit{De-Bias Net}s to capture the unmodeled part of the quadrotor dynamics. 
	
	We regard a sliding window of angular and linear velocity in the $\bodyframe$ and rotor speed data as input features to train the~\textit{Res-Dynamics Net} in a supervised fashion, whose history window size is set to 20.
	
	In order to fit the unmodeled force and fuse it into the EKF framework reasonably, a ${\text{MSE}}$ loss function is replaced by a Negative log-likelihood~(${\text{NLL}}$) loss~\cite{chen2021rnin} after the ${\text{MSE}}$ loss stabilizes and converges:
	\begin{equation}
	\begin{aligned}
		\mathcal{L}_{\text{MSE},\boldsymbol{f}}
		&= \frac{1}{N}\sum_{i=1}^N
		{\rVert 
			{^{\bodyframe}}{\accvel}{^{\graframe}_{\bodyframe}}_i-
			{^{\bodyframe}}\boldsymbol{\widehat{a}}{^{\graframe}_{\bodyframe}}_i
		\rVert^2_2},\\
		\mathcal{L}_{\text{NLL},\boldsymbol{f}}  
		&= \frac{1}{2N}\sum_{i=1}^N \log\det(\widehat{\Cov}^2_{\boldsymbol{f}_{i}}) \\
		&+ \frac{1}{2N}\sum_{i=1}^N
		{\rVert 
			{^{\bodyframe}}{\accvel}{^{\graframe}_{\bodyframe}}_i-
			{^{\bodyframe}}\boldsymbol{\widehat{a}}{^{\graframe}_{\bodyframe}}_i
		\rVert^2_{\widehat{\Cov}^2_{\boldsymbol{f}_{i}}}},
	\end{aligned}
	\end{equation}
	where
	\begin{equation}
	\begin{aligned}
	{^{\bodyframe}}\boldsymbol{\widehat{a}}{^{\graframe}_{\bodyframe}} &= 
	\frac{1}{m}
	(
	\tau U_{ss} {\boldsymbol{e}}_3 - 
	U_{s} D {_{\bodyframe}^{\graframe}}\mathbf{R}^\top 
	{{^{\graframe}}\boldsymbol{v}{_{\bodyframe}^{\graframe}}} +
	\boldsymbol{\widehat{f}}_{res}
	) - 
	{_{\bodyframe}^{\graframe}}\mathbf{R}^\top {^{\graframe}}\boldsymbol{\mathrm{g}},\\
	{^{\bodyframe}}{\accvel}{^{\graframe}_{\bodyframe}} &=
	{_{\bodyframe}^{\graframe}}\mathbf{R}^\top 
	{^{\graframe}}{\accvel}{^{\graframe}_{\imuframe}}
	+ {_{\bodyframe}^{\imuframe}}\mathbf{R}^\top 
	(
	{^{\imuframe}}{\widehat{\rotvel}} \times ({^{\imuframe}}{\widehat{\rotvel}} \times
	{^{\imuframe}}\boldsymbol{t}{^{\imuframe}_{\bodyframe}})
	+ {^{\imuframe}}\boldsymbol{{\boldsymbol{\widehat{\alpha}}}} \times 
	{^{\imuframe}}\boldsymbol{t}{^{\imuframe}_{\bodyframe}}
	).
	\end{aligned}
	\end{equation}
	Note that,
	$\boldsymbol{\widehat{f}}_{res}$ and
	$\widehat{\Cov}^2_{\boldsymbol{f}}$
	are the unmodeled force and its corresponding covariance,
	${^{\bodyframe}}\boldsymbol{\widehat{a}}{^{\graframe}_{\bodyframe}}$ is the intermediate variable calculated from $\boldsymbol{\widehat{f}}_{res}$,
	${^{\graframe}}{\accvel}{^{\graframe}_{\imuframe}}$ is the ground truth of acceleration,
	${^{\imuframe}}{\widehat{\rotvel}} = {^{\imuframe}}{\widetilde{\rotvel}}- {^{\imuframe}}\widehat{\bias}_{\rotvel}$,
	${^{\imuframe}}{\boldsymbol{\widehat{\alpha}}}=\frac{d}{dt}{^{\imuframe}}{\widehat{\rotvel}}$ and
	${^{\imuframe}}\boldsymbol{t}{^{\imuframe}_{\bodyframe}}$ is the extrinsic translation between the $\imuframe$ and $\bodyframe$ frame. 
	In practice, we low-pass filter the ${\widehat{\rotvel}}, {\boldsymbol{\widehat{\alpha}}}$ to reduce noise.
	
	\begin{figure}[t]
		\vspace{0.1cm}
		\begin{center}
			\includegraphics[angle=0,width=0.45\textwidth]{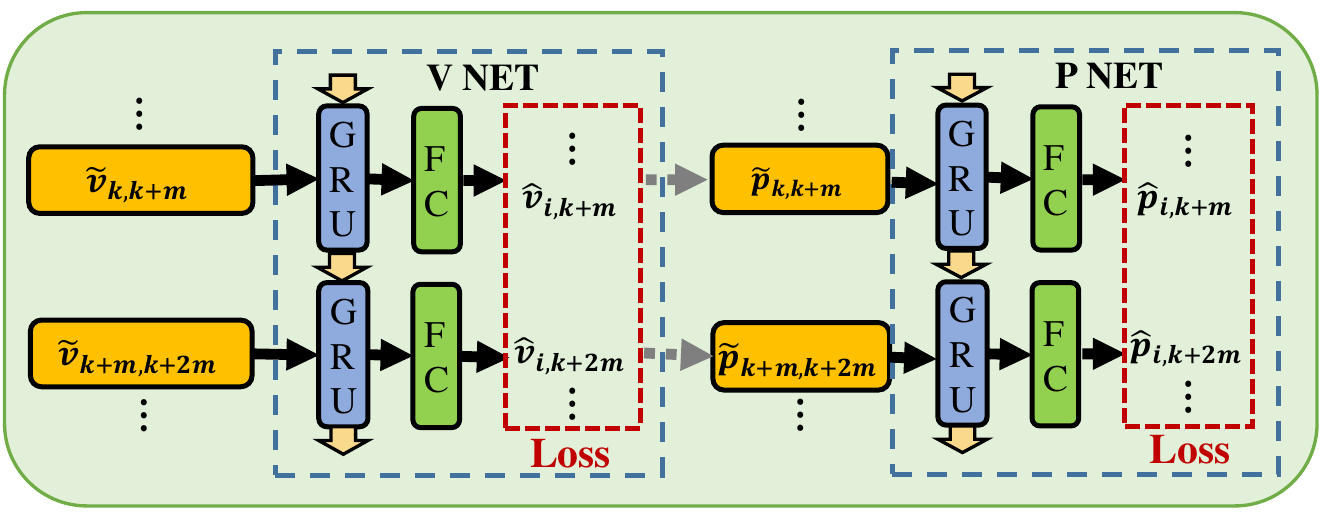}
			\caption{The network structure for~\textit{V-P Net}.}
			\label{vp_net}
		\end{center}
		\vspace{-1.2cm}
	\end{figure}
	\subsubsection{V-P Net}
	Even though~\textit{De-Bias Net}s help to reduce the bias of IMU measurements, 
	the remaining tiny offset of acceleration would not avoid the cumulative error in the integration process.
	Therefore, we design a cascaded GRU network to learn velocity and position approximation that has less error accumulation over a long period of time. 
	As every dimension of velocity and position are independent to each other, we adopt three cascaded GRU networks for the $X$, $Y$, and $Z$ axes separately.
	The cascaded architecture has two parts:~\textit{V Net} and~\textit{P Net}, these two parts have the same structure as shown in Fig.~\ref{vp_net}.
	
	Firstly, we integrate the de-biased acceleration as follow:
	\begin{equation}
	\begin{aligned}
	\widehat{\boldsymbol{v}}_{k,k+m} &= 
	\int_{k}^{k+m} 
	{_{\imuframe_t}^{\graframe}}\mathbf{R} ({^{\imuframe_t}}\widetilde{\accvel}-{^{\imuframe_t}}\widehat{\bias}_{\accvel}) dt, 
	\end{aligned}
	\end{equation}
	where the size of an integration window is $m$ without overlap among each window.
	Each axis of $\boldsymbol{\widehat{v}}_{k,k+m}$ and integration time $dt$ are input features of~\textit{V Net}.
	The hidden state is passed to fully-connected layers to regress the uniaxial velocity relative to the start of a sequence. 
	
	Then, we use results of the~\textit{V Net} to calculate the displacement over the integration window as shown in Eq.~(\ref{delta_p}), which is the input feature of~\textit{P Net} to regress the uniaxial position relative to the start of a sequence.
	\begin{equation}
	\label{delta_p}
	\begin{aligned}
	\boldsymbol{\widehat{p}}_{k,k+m} = 
	\boldsymbol{\widehat{v}}_{k+m}\boldsymbol{t}_{k,k+m} - 
	\iint_{k}^{k+m} 
	{_{\imuframe_t}^{\graframe}}\mathbf{R} ({^{\imuframe_t}}\widetilde{\accvel}-{^{\imuframe_t}}\widehat{\bias}_{\accvel})dt^{2}, 
	\end{aligned}
	\end{equation}
	where $	\boldsymbol{\widehat{v}}_{k+m} = \boldsymbol{\widehat{v}}_{i,k+m} + \boldsymbol{\widehat{v}}_{i}$,
	$\boldsymbol{\widehat{v}}_{i}$ is the initial velocity of the sequence, $ \boldsymbol{\widehat{v}}_{i,k+m} $ is the result of the~\textit{V Net} in the $(k+m)$-th step and $ \boldsymbol{t}_{k,k+m} $ is the duration of the integration window.
	
	Similarly, we use ${\text{MSE}}$ and ${\text{NLL}}$ loss over the whole sequence as follows:
	\begin{equation}
	\begin{aligned}
		\mathcal{L}_{\text{MSE},\boldsymbol{v},\boldsymbol{p}} &= 
		\frac{1}{2NM}\sum_{i=0}^N \sum_{j=0}^{M} 
		\left(
			\rVert 
				\boldsymbol{v}_{i,j} - \boldsymbol{\widehat{v}}_{i,j} 
			\rVert^2_2 + 
			\rVert 
				\boldsymbol{p}_{i,j} - \boldsymbol{\widehat{p}}_{i,j} 
			\rVert^2_2
		\right),\\
		\mathcal{L}_{\text{NLL},\boldsymbol{v},\boldsymbol{p}} &= 
		\frac{1}{2NM}\sum_{i=1}^N\sum_{j=0}^{M} 
		(\log\det(\widehat{\Cov}^2_{\boldsymbol{v}_{i,j}}) +
		 \log\det(\widehat{\Cov}^2_{\boldsymbol{p}_{i,j}}))\\ +
		\frac{1}{2NM} & \sum_{i=0}^N \sum_{j=0}^{M} 
		\big(
			\rVert 
			\boldsymbol{v}_{i,j} - \boldsymbol{\widehat{v}}_{i,j} 
			\rVert^2_{\widehat{\Cov}^2_{\boldsymbol{v}_{i,j}}} + 
			\rVert 
			\boldsymbol{p}_{i,j} - \boldsymbol{\widehat{p}}_{i,j} 
			\rVert^2_{\widehat{\Cov}^2_{\boldsymbol{p}_{i,j}}}
		\big),
	\end{aligned}
	\end{equation}
	where
	$\boldsymbol{v}_{i,j} =
	{^{\graframe}}\boldsymbol{v}{^{\graframe}_{\imuframe_j}}- {^{\graframe}}\boldsymbol{v}{^{\graframe}_{\imuframe_i}}$, 
	$\boldsymbol{p}_{i,j} =
	{^{\graframe}}\boldsymbol{p}{^{\graframe}_{\imuframe_j}}- {^{\graframe}}\boldsymbol{p}{^{\graframe}_{\imuframe_i}}$,
	$M$ is the sequence length, 
	$N$ is batch size, 
	$\boldsymbol{\widehat{v}}_{i,j}$ and $\boldsymbol{\widehat{p}}_{i,j}$ are the $j$-th predicted uniaxial velocity and position relative to the start of sequence $i$, 
	$\boldsymbol{v}_{i,j}$ and $\boldsymbol{p}_{i,j}$ are the ground truth,
	${\widehat{\Cov}^2_{\boldsymbol{v}_{i,j}}}$ and ${\widehat{\Cov}^2_{\boldsymbol{p}_{i,j}}}$ are the covariance of uniaxial velocity and position.
	

	
	

	\subsection{Dataset Preparation}
\begin{figure}[t]
		\vspace{0.3cm}
		\begin{center}
			\includegraphics[angle=0.8,width=0.36\textwidth]{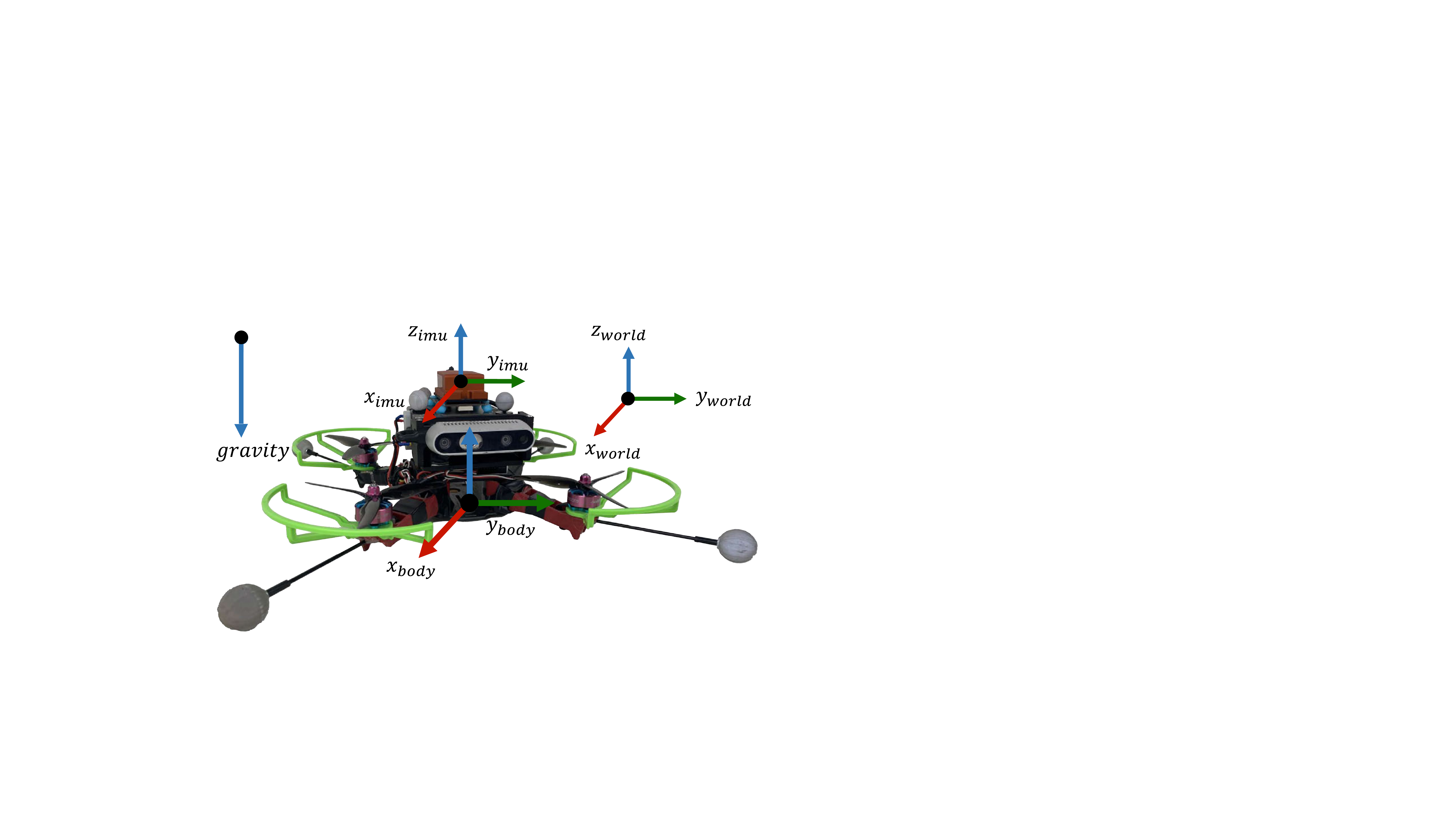}
			\caption{The quadrotor platform, which assembles an IMU and 4 tachometers, is built for data acquisition.}
			\label{assembly}
		\end{center}
		\vspace{-1.3cm}
	\end{figure}

	There are only the Blackbird dataset~\cite{antonini2020blackbird} and the VID dataset~\cite{zhang2021visual} in the flying robot community that contain quadrotor dynamics. 
	The former trajectories are too repetitive to generalize well for networks, while the latter suffers from insufficient data for training. 
	Therefore, to satisfy the training requirements, we equip a new quadrotor as a data acquisition platform and record adequate data. 
	The quadrotor assembles 4 tachometers and an additional IMU\footnote{https://www.xsens.com/mti-300} as shown in Fig.~\ref{assembly}, and the ground truth is provided by a VICON\footnote{https://www.vicon.com/} system.
	
	To ensure the diversity of trajectories, 
	we record a total of 274 yaw-constant and yaw-forward trajectories (total time: $325~min$, total distance: $ 19.92~km$), which include 247 Random, 15 Circle and 12 Figure-8 trajectories.
	Where Random trajectories are sampled and optimized~\cite{mellinger2011minimum} to be executed by the quadrotor as a smooth trajectory. 
	
	To provide high-frequency and accurate supervised data for network training,
	we first fit the full state of kinematics at the IMU frequency with B-splines~\cite{Geneva2020TRVICON2GT} in the gravity-aligned coordinate frame,
	and then identify the dynamics model of the quadrotor by offline optimization.

	\subsection{Training Details}
	\subsubsection{Covariance Training}
	Except for~\textit{De-Bias Net}, the mean and covariance outputs of the corresponding networks are available for the remaining three networks.
	Every covariance $\widehat{\Cov}^2$ is a 3$\times$3 matrix with 6 degrees of freedom, 
	but a diagonal form is assumed in this paper and parameterized by 3 coefficients written as
	~$
	\widehat{\Cov}^2_{\boldsymbol{\xi}} = 
	\text{diag}(
	e^{2\widehat{\boldsymbol{\xi}}_{x}}, 
	e^{2\widehat{\boldsymbol{\xi}}_{y}},
	e^{2\widehat{\boldsymbol{\xi}}_{z}}).
	$
	Since it is difficult to converge the direct training covariance by $\mathcal{L}_{\text{NLL}}$, we use $\mathcal{L}_{\text{MSE}}$ in the first 20 epoch and then replace by $\mathcal{L}_{\text{NLL}}$.
	\subsubsection{V-P Net}
	When training~\textit{V-P Net}, we use different lengths of sequences. 
	In experiment, the hidden sizes of GRU are [2, 64, 128, 256], integration window size $m$ is 20.
	\subsubsection{Data Separation and Optimizer}
	We select 200 Random trajectories for training and the remaining as the validation and test sets. Adam optimizer is chose to minimize the loss function, and the initial learning rate is $10^{-4}$. The length of validation and test set is from 58.5~($m$) to 107.1~($m$).
	
	\begin{table*}[t!]
		\vspace{0.2cm}
		\centering
		\begin{tabular}{cccccccc}
			\toprule
			& \makecell[c]{${^{\graframe}_{\imuframe}}\boldsymbol{q}$}
			& \makecell[c]{${^{\graframe}}\boldsymbol{p}{^{\graframe}_{\bodyframe}}$}
			& \makecell[c]{${^{\graframe}}\boldsymbol{v}{^{\graframe}_{\bodyframe}}$}
			& \makecell[c]{$\tau$}
			& \makecell[c]{$\boldsymbol{d}$}
			& \makecell[c]{${^{\imuframe}_{\bodyframe}}\boldsymbol{q}$}
			& \makecell[c]{${^{\imuframe}}\boldsymbol{t}{^{\imuframe}_{\bodyframe}}$} \\
			\toprule
			value   
			& ${^{\graframe}_{\imuframe_0}}\boldsymbol{q}$ 	
			& ${^{\graframe}}\boldsymbol{p}{^{\graframe}_{\imuframe_0}}$   
			& ${^{\graframe}}\boldsymbol{v}{^{\graframe}_{\imuframe_0}}$   
			& $1.1$
			& $[0,0,0]$   
			& $[1,0,0,0]$    
			& $[0,0,0]$	\\
			covariance  	
			& $1\mathrm{e}{-8} \times \mathbf{I}_3$   
			& $1\mathrm{e}{-4} \times \mathbf{I}_3$    	
			& $1\mathrm{e}{-6} \times \mathbf{I}_3$      
			& $1\mathrm{e}{-4}$    
			& $5\mathrm{e}{-4} \times \mathbf{I}_3$      	
			& $5\mathrm{e}{-5} \times \mathbf{I}_3$      
			& $5\mathrm{e}{-4} \times \mathbf{I}_3$      \\
			\bottomrule
		\end{tabular}
		\caption{The table lists initial values and covariances setting for all pose estimation experiments.}
		\label{intial_value_cov}
		\vspace{-0.6cm}
	\end{table*}
	
	\section{Inertial Dynamical Fusion Pipeline}
	\label{sec:Filter}
	\subsection{Rotation Stage} 
	\label{sec:rot-ekf}
	We separate the rotation and translation estimation in a two-stage EKF framework.
	\subsubsection{State}
	In the first rotation stage, there is only the rotation of the $\imuframe$ in the $\graframe$ frame taken as filter's state:
	\begin{equation}
	\boldsymbol{x} = {^{\graframe}_{\imuframe}}\boldsymbol{q}.
	\end{equation}
	
	\subsubsection{Process Model}
	The rotational equation is given:
	\begin{equation}
	\label{process_rot}
	\begin{aligned}
	\boldsymbol{\dot{x}} 
	= \frac{1}{2} {^{\graframe}_{\imuframe}}\boldsymbol{q} \otimes {^{\imuframe}}{\rotvel} 
	= \frac{1}{2} {^{\graframe}_{\imuframe}}\boldsymbol{q} \otimes ({^{\imuframe}}{\widehat{\rotvel}}+{\ngvel}),
	\end{aligned}
	\end{equation}
	where
	${^{\imuframe}}{\widehat{\rotvel}}=
	{^{\imuframe}}{\widetilde{\rotvel}}-
	{^{\imuframe}}\widehat{\bias}_{\rotvel}$, 
	${\ngvel} \sim \mathcal{N}(0,\Cov_{\rotvel}^2)$, and
	${^{\imuframe}}\widehat{\bias}_{\rotvel}$ is the output of the gyroscope~\textit{De-Bias Net}.
	
	\subsubsection{Measurement Model}
	The attitude controllers of most flying robots rely on the complementary filters~\cite{mahony2008nonlinear,madgwick2011estimation} to obtain attitude observations.
	Similarly, we consider the gravity alignment constraint to obtain the tilt observation:
	\begin{equation}
	\begin{aligned}
	\label{gravity_obs}
	^{\imuframe}\widehat{\accvel} \approx 
	{_{\imuframe}^{\graframe}}\mathbf{R} ^\top 
	{^{\graframe}}\boldsymbol{\mathrm{g}} + {\navel}.
	\end{aligned}
	\end{equation}
	Similarly, ${^{\imuframe}}{\widehat{\accvel}}=
	{^{\imuframe}}{\widetilde{\accvel}}-
	{^{\imuframe}}\widehat{\bias}_{\accvel}$, 
	${\navel} \sim \mathcal{N}(0,\Cov_{\accvel}^2)$, and
	${^{\imuframe}}\widehat{\bias}_{\accvel}$ is the output of the accelerator~\textit{De-Bias Net}.
	

	\subsection{Translation Stage} 
	\label{sec:tra-ekf}
	\subsubsection{State}
	The state of the second stage is defined as:
	\begin{equation}
	\boldsymbol{x} = ({^{\graframe}}\boldsymbol{p}{^{\graframe}_{\bodyframe}},~
	{^{\graframe}}\boldsymbol{v}{^{\graframe}_{\bodyframe}},~
	\tau,~
	\boldsymbol{d},~
	{^{\imuframe}_{\bodyframe}}\boldsymbol{q},~
	{^{\imuframe}}\boldsymbol{t}{^{\imuframe}_{\bodyframe}}),
	\end{equation}
	where
	${^\graframe}\boldsymbol{p}{^{\graframe}_{\bodyframe}}$ and 
	${^\graframe}\boldsymbol{v}{^{\graframe}_{\bodyframe}}$ are respectively the velocity and position of the quadrotor body $\bodyframe$ frame expressed in the
	$\graframe$ frame,
	$\boldsymbol{d}$ is the drag vector of $\left(d_x, d_y, d_z\right)$, and
	$({^{\imuframe}_{\bodyframe}}\boldsymbol{q},{^{\imuframe}}\boldsymbol{t}{^{\imuframe}_{\bodyframe}})$ is the extrinsic parameter between the $\bodyframe$ and the $\imuframe$ frame.
	\subsubsection{Process Model}
	We regard the quadrotor dynamics as the input, and express the complete process model as follows:
	\begin{equation}
	\label{process_tra}
	\begin{aligned}
	{^{\graframe}}\boldsymbol{\dot{p}}{^{\graframe}_{\bodyframe}} &=
	{^{\graframe}}\boldsymbol{v}{^{\graframe}_{\bodyframe}}, \\
	{^{\graframe}}\boldsymbol{\dot{v}}{^{\graframe}_{\bodyframe}} &= 
	\frac{1}{m} {_{\bodyframe}^{\graframe}}\mathbf{R} 
	( 
	\tau U_{ss} \boldsymbol{e}_3-
	U_{s} D {_{\bodyframe}^{\graframe}}\mathbf{R}^\top
	{^{\graframe}}\boldsymbol{v}{^{\graframe}_{\bodyframe}} +
	\widehat{\boldsymbol{f}}_{res} + \nb_{\boldsymbol{f}}
	) 
	-{^{\graframe}}\boldsymbol{\mathrm{g}}, \\
	\dot{k}_t &= 0, \quad
	\dot{\boldsymbol{d}} = \boldsymbol{0}, \quad
	{^{\imuframe}_{\bodyframe}}\boldsymbol{\dot{q}} = \boldsymbol{0}, \quad
	{^{\imuframe}}\boldsymbol{\dot{t}}{^{\imuframe}_{\bodyframe}} = \boldsymbol{0},
	\end{aligned}
	\end{equation}
	where ${_{\bodyframe}^{\graframe}}\mathbf{R} = 
	{_{\imuframe}^{\graframe}}\mathbf{R} {_{\bodyframe}^{\imuframe}}\mathbf{R}$, and $\widehat{\boldsymbol{f}}_{res}$ and $\widehat{\Cov}_{\boldsymbol{f}}^2$
	$(\nb_{\boldsymbol{f}} \sim \mathcal{N}(0,\widehat{\Cov}_{\boldsymbol{f}}^2))$ 
	are the~\textit{Res-Dynamics Net} outputs.

	\subsubsection{Measurement Model}
	Firstly, since accelerometer measurements are not used before, we take the inconsistency of the ${\imuframe}$ and ${\bodyframe}$ frame into account and express the dynamics measurement equation as follow:
	\begin{equation}
	\label{acc_obs}
	\begin{aligned}
	^{\imuframe}\widehat{\accvel}
	&= \frac{1}{m} {_{\bodyframe}^{\imuframe}}\mathbf{R} 
	\left( 
	\tau U_{ss} \boldsymbol{e}_3-
	U_{s} D {_{\bodyframe}^{\graframe}}\mathbf{R}^\top 
	{^{\graframe}}\boldsymbol{v}{^{\graframe}_{\bodyframe}} +
	\widehat{\boldsymbol{f}}_{res}
	\right) \\
	& - {^{\imuframe}}{\widehat{\rotvel}} \times ({^{\imuframe}}{\widehat{\rotvel}} \times
	{^{\imuframe}}\boldsymbol{t}{^{\imuframe}_{\bodyframe}})
	- {^{\imuframe}}{\widehat{\boldsymbol{\alpha}}} \times 
	{^{\imuframe}}\boldsymbol{t}{^{\imuframe}_{\bodyframe}}
	+ \nb_{a}.
	\end{aligned}
	\end{equation}
	What's more, our aforementioned~\textit{V-P Net} can be used as observers of velocity and displacement:
	\begin{equation}
	\label{vp_obs}
	\begin{aligned}
	{^{\graframe}}\boldsymbol{\widehat{v}}{^{\graframe}_{\imuframe}} & =
	{^{\graframe}}\boldsymbol{v}{^{\graframe}_{\bodyframe}} -
	{_{\imuframe}^{\graframe}}\mathbf{R}
	\left(
	{^{\imuframe}}{\widehat{\rotvel}} \times {^{\imuframe}}\boldsymbol{t}{^{\imuframe}_{\bodyframe}}
	\right) +
	\nb_{v},\\
	{^{\graframe}}\boldsymbol{\widehat{p}}{^{\graframe}_{\imuframe}} & = 
	{^{\graframe}}\boldsymbol{p}{^{\graframe}_{\bodyframe}}
	-{_{\imuframe}^{\graframe}}\mathbf{R}
	{^{\imuframe}}\boldsymbol{t}{^{\imuframe}_{\bodyframe}}
	+\nb_{p}.
	\end{aligned}
	\end{equation}  
	Specially,~\textit{V-P Net} outputs
	${^{\graframe}}\boldsymbol{\widehat{v}}{^{\graframe}_{\imuframe}}$ and
	${^{\graframe}}\boldsymbol{\widehat{p}}{^{\graframe}_{\imuframe}}$ as well as their covariances
	$\widehat{\Cov}_{\boldsymbol{v}}^2$ 
	$(\nb_{\boldsymbol{v}} \sim \mathcal{N}(0,\widehat{\Cov}_{\boldsymbol{v}}^2))$
	and
	$\widehat{\Cov}_{\boldsymbol{p}}^2$ 
	$(\nb_{\boldsymbol{p}} \sim \mathcal{N}(0,\widehat{\Cov}_{\boldsymbol{p}}^2))$.
	More details about the EKF process and observability proof can be found in the supplementary material~\cite{suppmat}.

	\section{Experiments}
	\label{sec:Experiments}
	\subsection{Metrics Definition}
	In order to assess the performance of our system, we define the metrics including
	the Absolute Translation Error~($\textbf{ATE}$),
	the Absolute Rotation Error~($\textbf{ARE}$),
	the Relative Translation Error~($\textbf{RTE}$),
	the Relative Rotation Error~($\textbf{RRE}$), and
	the Translation Drift~($\textbf{TD}$),
	the Rotation Drift~($\textbf{RD}$) as follows:
	\begin{itemize}
		\label{metric}
		\item{\textbf{ATE}}~(\textrm{m})$:= \sqrt{\frac{1}{l}\sum_{i=1}^l 
			\|{^{\graframe}}\boldsymbol{p}_i -
			{^{\graframe}}\boldsymbol{\widehat{p}}_i
			\|^2_2}$,\\
		\item{\textbf{ARE}}~(\textrm{m})$:= \sqrt{\frac{1}{l}\sum_{i=1}^l
			\|\boldsymbol{d}
			({^{\graframe}_{\imuframe_i}}\boldsymbol{q},
			{^{\graframe}_{\imuframe_i}}\boldsymbol{\widehat{q}})
			\|^2_2}$,\\
		\item{\textbf{RTE}}~(\%)$:= \sqrt{\frac{1}{l}\sum_{i=1}^l
			\|{^{\graframe}}\boldsymbol{p}_{i,i+\Delta t}-
			{^{\graframe}}\boldsymbol{\widehat{p}}_{i,i+\Delta t}
			\|^2_2}$, \\
		\item{\textbf{RRE}}~(\textrm{rad})$:= \sqrt{\frac{1}{l}\sum_{i=1}^l
			\|\boldsymbol{d}
			({^{\imuframe_{i+\Delta t}}_{\imuframe_i}}\boldsymbol{q},
			{^{\imuframe_{i+\Delta t}}_{\imuframe_i}}\boldsymbol{\widehat{q}})
			\|^2_2}$, \\
		\item{\textbf{TD}}~(\textrm{rad})$:= 
		\|{^{\graframe}}\boldsymbol{p}_l-
		{^{\graframe}}\boldsymbol{\widehat{p}}_l
		\|^2_2 /(\text{trajectory-length})$,\\
		\item{\textbf{RD}}~($\textrm{rad}/\min$)$:= 
		\|\boldsymbol{d}
		({^{\graframe}_{\imuframe_l}}\boldsymbol{q},
		{^{\graframe}_{\imuframe_l}}\boldsymbol{\widehat{q}})
		\|^2_2 /(\text{sequence-duration})$,
	\end{itemize}
	where the rotation distance is
	$\boldsymbol{d}(\boldsymbol{q},\boldsymbol{\widehat{q}})$ = 
	$log(conj(\boldsymbol{\widehat{q}}) \otimes \boldsymbol{q})$, 
	$l$ is the length of each dataset and $\Delta t$ is set to $1/20s$.

	\subsection{Pose Comparison}
	To better demonstrate our system, we select several conventional and learning-based algorithms to compare the estimation accuracy of 6D pose in the test set. Note that we set the initial values and covariances in TABLE.\ref{intial_value_cov}.
	The proposed DIDO and TLIO~\cite{liu2020tlio} are the complete systems whose results are the filter output.
	We train and test the former RoNIN-ResNet~\cite{yan2019ronin} by providing the ground truth rotation, so there is no rotation error shown.
	There are two conventional algorithms, Dyn-UKF~\cite{svacha2019inertial} and Ori-EKF~\cite{sabatelli2012double}.
	Dyn-UKF~\cite{svacha2019inertial}, integrates the dynamics as a forward process and updates the tilt and velocity in ${\bodyframe}$ frame by the raw accelerometer measurements, while all states are converted to the ${\graframe}$ frame by estimated rotation.
	Ori-EKF~\cite{sabatelli2012double} only obtains the rotation using the raw IMU measurements.
	
\begin{figure}[t!]
		\vspace{0.2cm}
		\begin{center}
			\includegraphics[angle=0,width=0.435\textwidth]{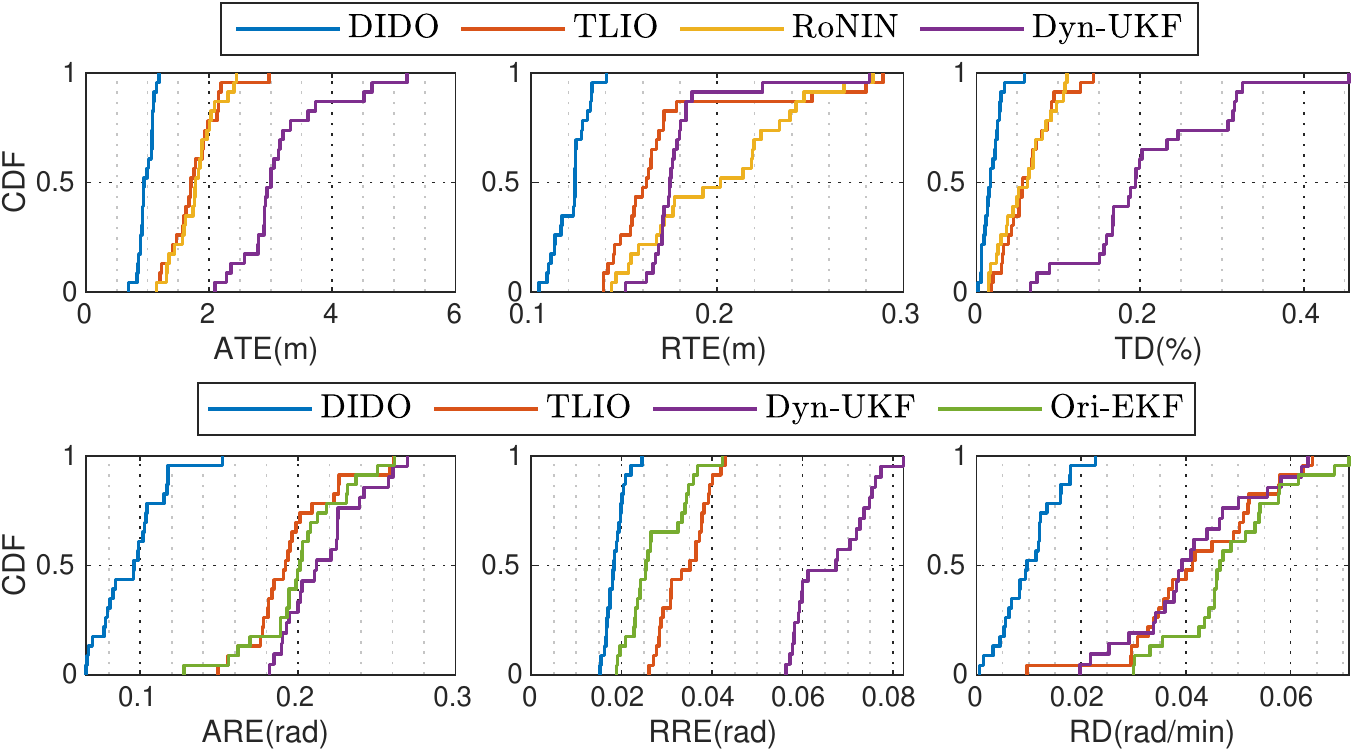}
			\caption{
				Pose comparison between DIDO and other methods. 
				Each subplot shows the cumulative density function (CDF) on the entire test set. The closer the graph is to the left, the better the performance.}
			\label{cdf}
		\end{center}
		\vspace{-1.5cm}
	\end{figure}

\begin{figure}[t!]
		\vspace{-0.0cm}
		\begin{center}
			\includegraphics[angle=0,width=0.445\textwidth]{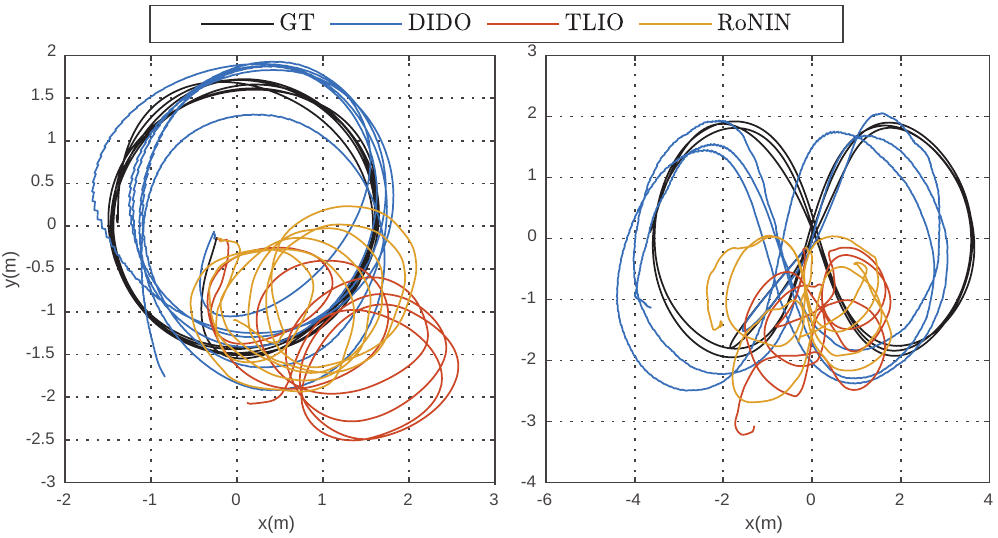}
			\caption{
				Top view diagrams of on two test trajectories.}
			\label{xoy}
		\end{center}
		\vspace{-1.7cm}
	\end{figure}

\begin{figure*}[t!]
		\vspace{0.0cm}
		\begin{center}
			\includegraphics[angle=0,width=0.80\textwidth]{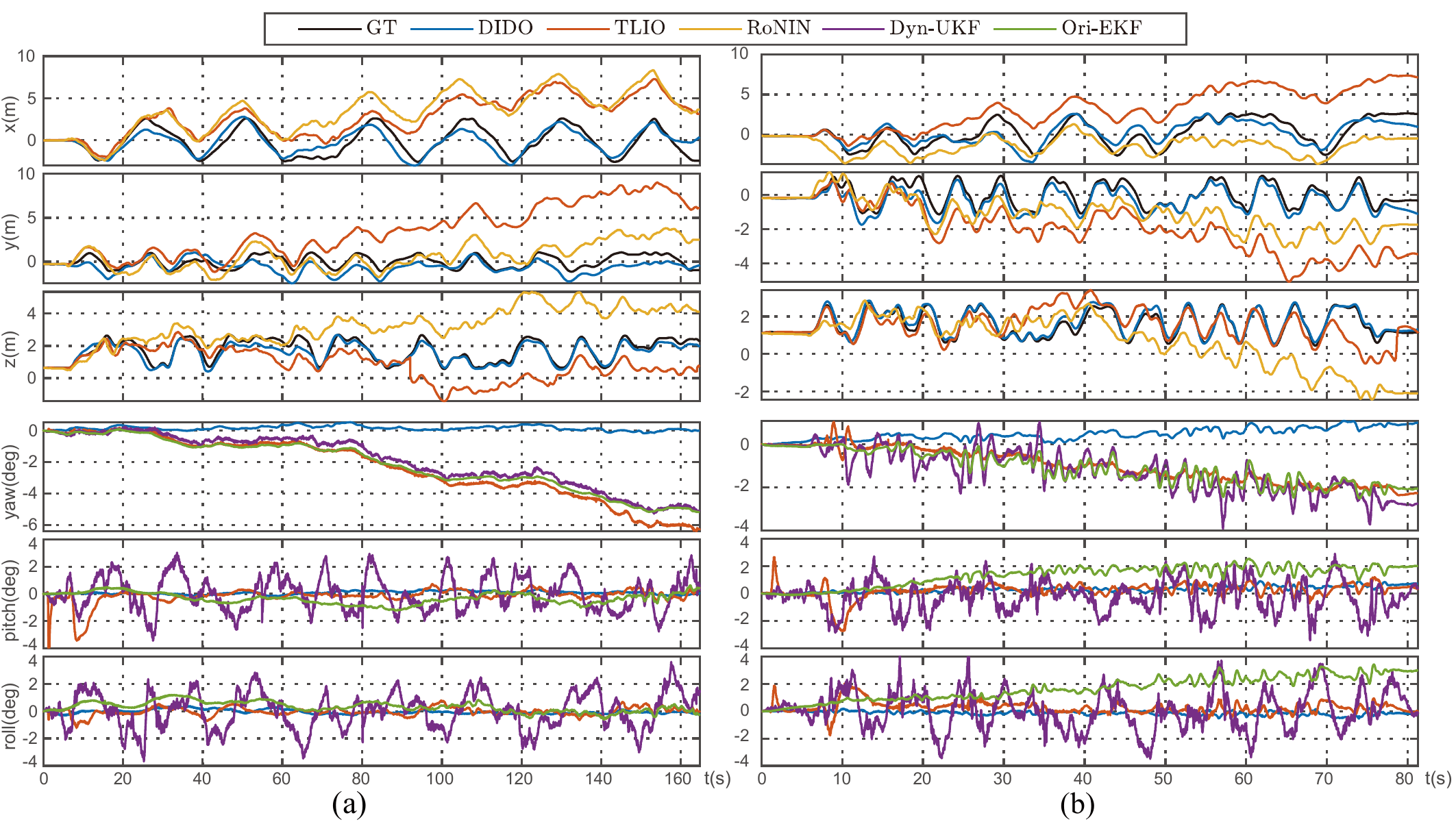}
			\caption{Demonstration of different methods for pose estimation. 
				There are two Random trajectories corresponding to different max uniaxial velocities (a:0.5m/s, b:2.5m/s) in the test set.
				The ground truth positions are plotted as black lines in the first row while ignoring the Dyn-UKF~\cite{svacha2019inertial} due to its divergence. 
				The rotation errors are drawn in the second row.}
			\label{pose}
		\end{center}
		\vspace{-0.5cm}
	\end{figure*}

	Fig.~\ref{cdf} shows the error distributions among the entire test set, which distinctly shows the proposed DIDO outperforms other methods.
	Dyn-UKF~\cite{svacha2019inertial} based only on the momentum theory does not work well, while DIDO compensates for the biased dynamics model resulting in a better performance.
	Fig.~\ref{xoy} presents three learning-based methods for position estimation on unseen simple trajectories.
	Fig.~\ref{pose} depicts the performance of different methods for each axis.
	TLIO~\cite{liu2020tlio} and RoNIN-ResNet~\cite{yan2019ronin} both have unavoidable cumulative drift, while attributing to the~\textit{V-P Net}, DIDO has stable position estimation. 
	Besides, \textit{De-Bias Net} decelerates the severe yaw drift, which effectively helps the position estimation.

	\subsection{Parameters Estimation}

\begin{figure}[t!]
		\vspace{-0.1cm}
		\begin{center}
			\includegraphics[angle=0,width=0.445\textwidth]{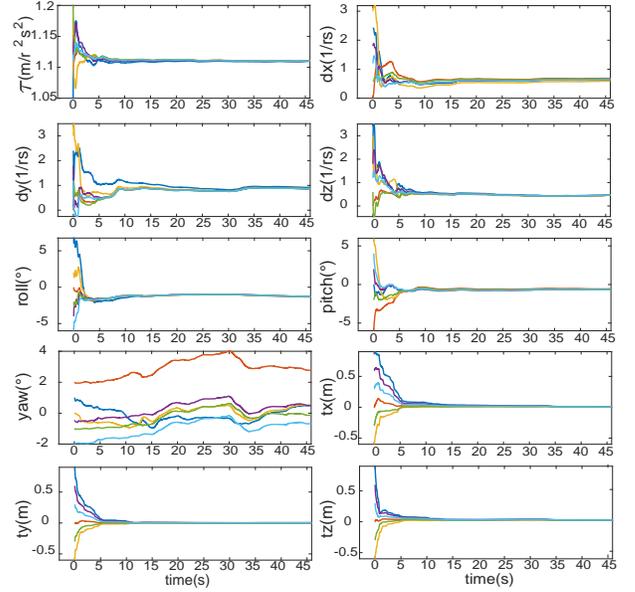}
			\caption{Parameters convergence for 6 runs.}
			\label{estimated_param}
		\end{center}
		\vspace{-0.5cm}
	\end{figure}

	To show the effectiveness and robustness of the parameter identification performance of the proposed system, we conduct an experiment by perturbing the initial values of the estimated parameters.
	Fig.~\ref{estimated_param} presents the convergence of the parameters in six independent runs, in which the majority of the parameters can converge quickly, except for the extrinsic parameter rotation expressed in terms of Euler angles ($XYZ$ in Tait–Bryan angles).
	This is due to the fact that the quadrotor in our dataset does not fly fast enough, so the yaw angle of the extrinsic rotation appears to be weakly observable. The other degradation cases are:
	when
	${^{\imuframe}}{\widehat{\rotvel}}$ = $\boldsymbol{0}_{3\times1}$,
	${^{\imuframe}}\boldsymbol{t}{^{\imuframe}_{\bodyframe}}$ is unobservable, and
	when 
	${^{\graframe}}\boldsymbol{v}{^{\graframe}_{\bodyframe}}$ = $[0;0;*]$,
	$D$ and ${_{\bodyframe}^{\imuframe}}\mathbf{R}$ are unobservable.
	The detailed proof process is provided in the supplementary material~\cite{suppmat}.
	
	\section{Ablation study}
	\label{sec:Ablations}
	To be able to better illustrate the effectiveness of the proposed system, we verify the behavior of each network module through ablation experiments.
	\subsection{De-Bias Net}
\begin{figure}[t!]
		\vspace{0cm}
		\begin{center}
			\includegraphics[angle=0,width=0.435\textwidth]{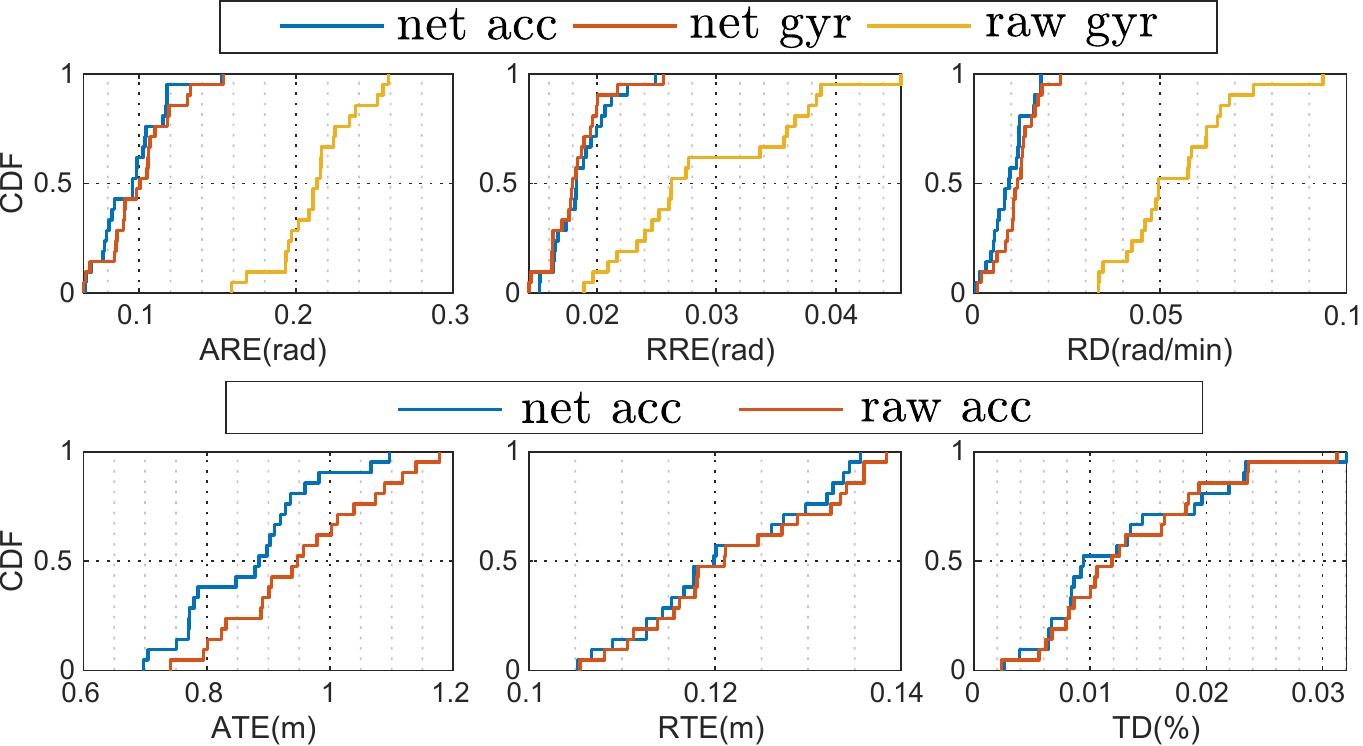}
			\caption{{De-Bias Net} ablation study.
				The first row shows the rotational comparison between the three methods
				\textit{net acc} (debiased gyroscope
				${\widehat{\rotvel}}$ = ${\widetilde{\rotvel}}$ - $\widehat{\bias}_{\rotvel}$ integrating and debiased accelerometer
				${\widehat{\accvel}}$ = ${\widetilde{\accvel}}$ - $\widehat{\bias}_{\accvel}$ updating),
				\textit{net gyr} (${\widehat{\rotvel}}$ integrating) and
				\textit{raw gyr} (${\widetilde{\rotvel}}$ integrating).
				The second row shows the translational comparison between 
				\textit{net acc} (with accelerometer~\textit{De-Bias Net}) and 
				\textit{raw acc} (without accelerometer~\textit{De-Bias Net}).}
			\label{debias_net}
		\end{center}
		\vspace{-1.5cm}
	\end{figure}

	We compare the rotation metrics for the gyroscope~\textit{De-Bias Net} and translation metrics for the accelerometer one, respectively.
	As shown in the Fig.~\ref{debias_net},
	the debiased gyroscope significantly outperforms method without debiasing, and updating by the debiased accelerometer results in better absolute rotation but with a slight loss of relative rotation.
	The debiased accelerometer also contributes to translation because debiasing facilitates the estimation of velocity increment (the input of~\textit{V-P Net}) and thus improves the accuracy of the~\textit{V-P Net} observation. Note that, the system is updated by Eq.(\ref{acc_obs}) and Eq.(\ref{vp_obs}) when verifying the accelerometer~\textit{De-Bias Net}.
	

	\subsection{Res-Dynamics Net}
\begin{figure}[t!]
		\vspace{0.1cm}
		\begin{center}
			\includegraphics[angle=0,width=0.445\textwidth]{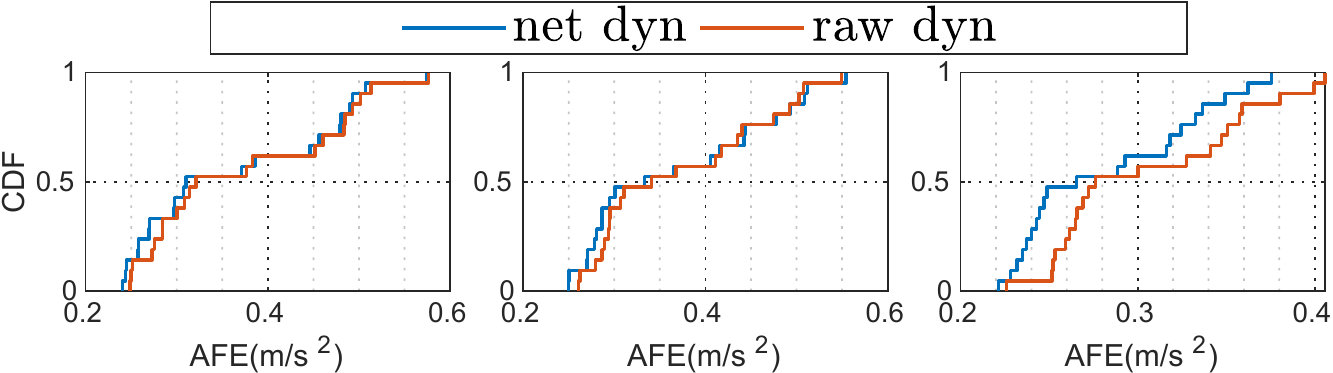}
			\caption{{Res-Dynamics Net} ablation study.
				The metrics in three directions are used to show the force estimation of
				\textit{net dyn} and \textit{raw dyn}
				(with and without \textit{Res-Dynamics Net}).}
			\label{f_net_dyn}
		\end{center}
		\vspace{-0.5cm}
	\end{figure}

	The~\textit{Res-Dynamics Net} depends on dynamical parameters and other motion states which need to be estimated online in the proposed system, so we validate the tri-axial forces by comparing the complete EKF process with and without the~\textit{Res-Dynamics Net}.
	Similarly, we also define a metric to measure the precision of force estimation:
	\begin{itemize}
		\label{f_metric}
		\item{\textbf{AFE}}$:= \sqrt{\frac{1}{l}\sum_{i=1}^l 
			\|\boldsymbol{f}_i - \boldsymbol{\widehat{f}}_i\|^2_2}$.
	\end{itemize}
	As shown in the Fig.~\ref{f_net_dyn},~\textit{Res-Dynamics Net} allows for more accurate dynamics modeling, especially in the z-axis.

	\subsection{V-P Net}
\begin{figure}[t!]
		\vspace{-0.00cm}
		\begin{center}
			\includegraphics[angle=0,width=0.435\textwidth]{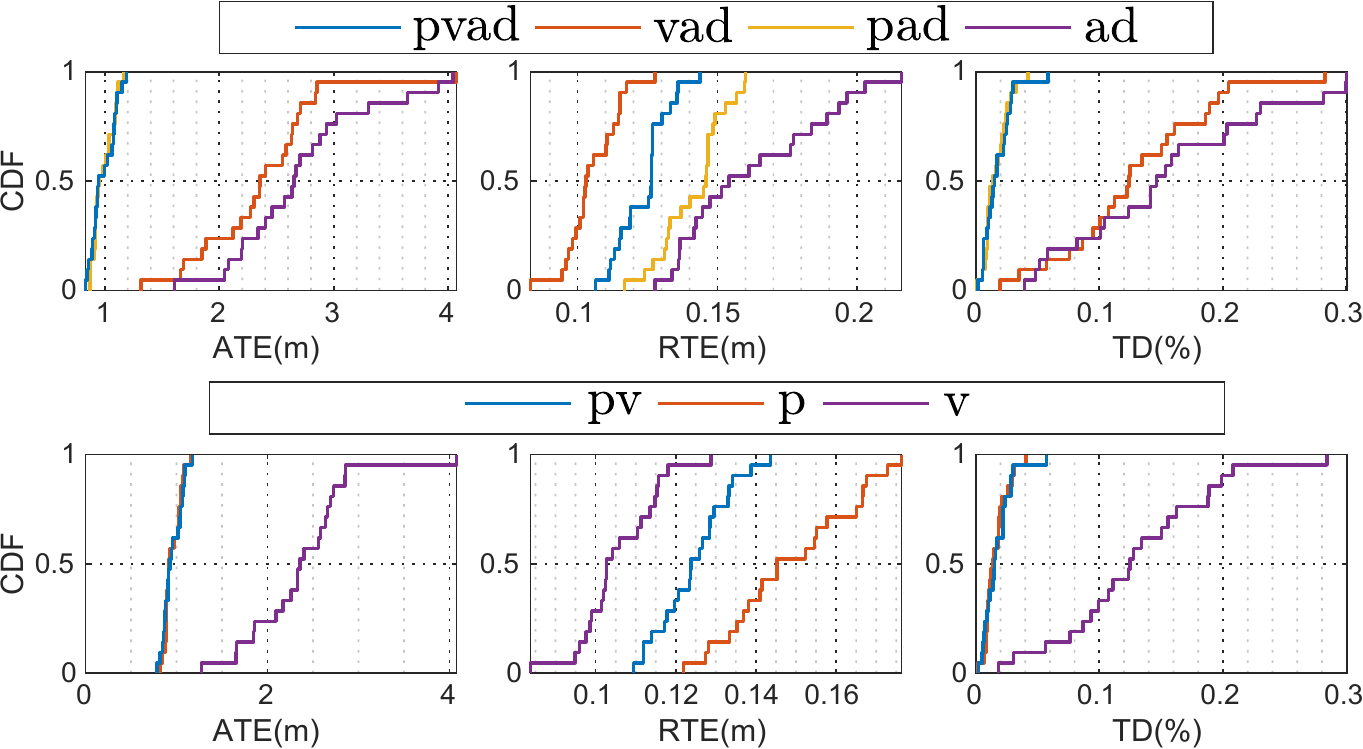}
			\caption{{V-P Net} ablation study.
				Translational estimation is compared under six different combinations of 
				\textit{p} (position update),
				\textit{v} (velocity update),
				\textit{a} (acceleration update) and
				\textit{d} (using the \textit{Res-Dynamics Net}).}
			\label{p_vp_net}
		\end{center}
		\vspace{-1.7cm}
	\end{figure}

	To verify the effectiveness of the~\textit{V-P Net}, we design two sets of experiments.
	As shown in the Fig.~\ref{p_vp_net},~\textit{P Net} and~\textit{V Net} enhance the estimation accuracy of position and velocity, respectively.
	Furthermore, using~\textit{V-P Net} results in better RTE with guaranteed ATE in translation estimation.
	
	\subsection{One-stage or Two-stage}
\begin{figure*}[h!]
		\vspace{0.0cm}
		\begin{center}
			\includegraphics[angle=0,width=0.80\textwidth]{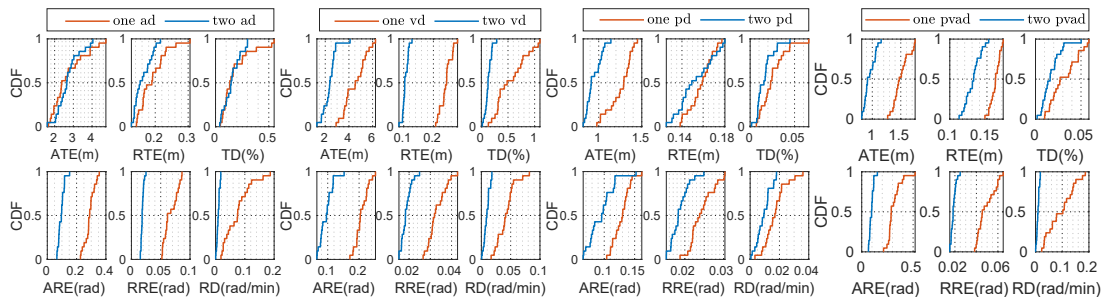}
			\caption{Pose Comparison of EKF. 
				One-stage and two-stage EKF are abbreviated as \textit{one} and \textit{two}, respectively.	\textit{p}, \textit{v}, \textit{a} and \textit{d} hold the same meanings as Fig.~\ref{p_vp_net}. For example, \textit{one vd} refers to a one-stage EKF updated by velocity in the case of the \textit{Res-Dynamics Net}.
			} 
			\label{one_two}
		\end{center}
		\vspace{-0.8cm}
	\end{figure*}

	To demonstrate the effect of these three measurement models on the state estimation system, we design four sets of experiments.
	As shown in the Fig.~\ref{one_two},
	rotation is prone to be incorrectly updated by each noisy observation in the one-stage EKF, which further affects the estimation of velocity and position.
	This is due to the fact that rotation is very sensitive as an explicit or implicit input to the networks. 
	Once the rotation can not be updated correctly during the EKF process, it will make the outputs of networks worse and thus impair the state estimation system.

	\section{Discussions}
	\label{sec:Discussions}
	\subsection{Dynamics or IMU based}
	In this sensor fusion system, there are two ways to obtain acceleration, 
	one is the accelerometer measurements, and the other is the quadrotor dynamics.
	However, if the accelerometer measurements are used as the process model, the constraint between tilt and velocity does not hold.
	As a consequence, the dynamics-based methods may be superior to the IMU-based one in the quadrotor estimation system.
	
	\subsection{Covariance}
	
	Covariances are of great importance because they can be used to describe the data distribution and also taken as weights for multi-sensor fusion.
	The neural network covariances can dynamically depict the uncertainty of the network output mean values, which reduce the inaccuracy led by the static covariance setting. However, in the EKF process, the covariances may propagate inaccurately due to the linear approximation of the process and measurement models. Therefore, we do not directly use the neural network covariances but scale the covariances in a similar way to TLIO to obtain a better estimation.
	
	\section{Conclusions and Future work}
	\label{sec:Conclusions}
	\subsection{Conclusions}
	In this work, we propose an inertial and quadrotor dynamical odometry system introducing deep neural networks in a two-stage tightly-coupled EKF framework.
	To the best of our knowledge, this is the first estimation framework that combines IMU, quadrotor dynamics and deep learning methods to simultaneously estimate kinematic, and dynamic states, as well as extrinsic parameters.
	To take full advantage of the two interoceptive sensors, we design deep neural networks to regress the IMU bias, the dynamics compensation, motion states of the quadrotor, as well as their covariances.
	
	Experimental results have demonstrated that the proposed method outperforms the other conventional and learning-based methods in pose estimation.
	It is accurate and invulnerable to consider~\textit{De-Bias Net} and gravity alignment constraints for rotation estimation.
	And it is efficient and versatile to take into account both quadrotor dynamics and deep neural networks for translation estimation.
	The code and data will be open-sourced to the community~\footnote{https://zhangkunyi.github.io/DIDO/}.
	
	\subsection{Future work}
	\label{sec:Future work}
	Even though the proposed method makes use of deep neural networks to improve the accuracy of estimation, it still does not guarantee a perfect globally consistent yaw angle and position under long and highly maneuverable flight.
	The plausible solution is to provide an absolute observation, such as GPS or magnetic compass.
	Besides, some of the proposed modules can be supplied in GPS-free environments with exteroceptive sensors to improve the robustness of estimation. These additional sensors can be fused with the proposed method by means of optimization or filtering.

	
	\newlength{\bibitemsep}\setlength{\bibitemsep}{0\baselineskip}
	\newlength{\bibparskip}\setlength{\bibparskip}{0pt}
	\let\oldthebibliography\thebibliography
	\renewcommand\thebibliography[1]{%
		\oldthebibliography{#1}%
		\setlength{\parskip}{\bibitemsep}%
		\setlength{\itemsep}{\bibparskip}%
	}

	\bibliography{IROS2022_KyZhang}
	
	\begin{appendices} 
	\end{appendices}

\end{document}